\documentclass{article}
\usepackage{spconf,amsmath,graphicx}
\usepackage{url}

\usepackage{enumitem}
\setlist{nosep, leftmargin=14pt}

\usepackage{mwe} 
\usepackage{amssymb} 
\usepackage{bm} 
\usepackage{color}

\title{An Ordinal Diffusion Model for Generating Medical Images with Different Severity Levels}

\name{Shumpei Takezaki \qquad Seiichi Uchida\vspace{-2mm}}
\address{Kyushu University, Fukuoka, Japan \vspace{-2mm}}
\begin{document}
%
\maketitle
\begin{abstract}
Diffusion models have recently been used for medical image generation because of their high image quality. In this study, we focus on generating medical images with {\em ordinal classes}, which have ordinal relationships, such as severity levels. We propose an {\em Ordinal Diffusion Model} (ODM) that controls the ordinal relationships of the estimated noise images among the classes. Our model was evaluated experimentally by generating retinal and endoscopic images of multiple severity classes. ODM achieved higher performance than conventional generative models by generating realistic images, especially in high-severity classes with fewer training samples.
\end{abstract}
%
\section{Introduction}
\label{sec:intro}
Recent trials on medical image generation use diffusion models~\cite{ddpm} because of their high-quality results. Diffusion models comprise two iterative processes: a forward process and a reverse process. The forward process transforms a real image $\bm{x}$ into a random noise image by iteratively adding a noise $\bm{\epsilon}$. The reverse process is called a ``denoising'' process to iteratively estimate the noise image $\bm{\epsilon}$ added to $\bm{x}$. The denoising process is realized by training a neural network model (especially U-Net). Using the trained model, generating various realistic images from noise images is possible.
\par 

\begin{figure}[t]
    \centering
    \includegraphics[width=0.9\linewidth]{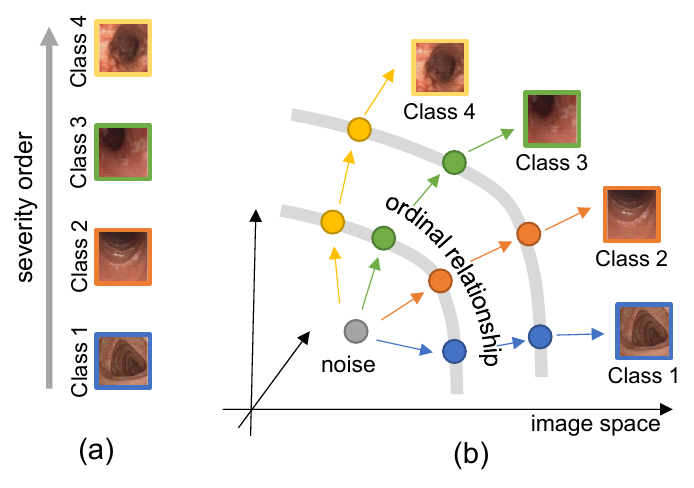}\\[-2mm]
        \caption{(a)~Endoscopic images with four ordinal severity classes of UC.
(b) An overview of the proposed ordinal diffusion model. Its denoising process generates images while reflecting ordinal class relationships.}
    \label{fig:overview}
\end{figure}

This paper focuses on generating medical images with {\em ordinal classes}.
For example, when the classes $\mathcal{C}=\{1, \ldots, C\}$ are ordinal, the images from the classes $c$ and $c'$ ($c<c')$ have some ordinal relationship. 
Ordinal classes are common in various CV/PR tasks, such as ages~\cite{age_person}, facial expressions~\cite{zhao_2016_CVPR}, and ratings~\cite{movie_star}.
A typical example of ordinal classes of medical images is severity levels. Fig.~\ref{fig:overview}~(a) shows the ordinal severity 
classes (called Mayo~\cite{uc}) of ulcerative colitis~(UC) endoscopic images.
\par 

Our key idea is that the ordinal relationships will be useful for estimating class distributions and, therefore, for generating images of the ordinal classes.
For example, for the neighboring ordinal classes $c-1$, $c$, and $c+1$, the samples from $c-1$ and $c+1$ help estimate the distribution of the class $c$ by their interpolation effect. Similarly, the samples of $c-1$ and $c$ will help estimate the distribution of $c+1$ by extrapolation. The extrapolation effect will be especially useful in the higher severity classes, where sample sizes are often limited. Note that standard diffusion models conditioned by each class $c$ cannot expect these interpolation and extrapolation effects.
\par 
This paper proposes an {\em Ordinal Diffusion Model} (ODM) for generating images with ordinal classes. Fig.~\ref{fig:overview}~(b) shows an overview of the proposed model. 
Unlike standard diffusion models, ODM can treat the ordinal relationship among classes. Specifically, by introducing a new loss function, ODM controls the estimated {\em noise images} to have ordinal relationships among the classes. In other words, we try to prove that regulating noise images is beneficial for regulating the final generated images to keep their ordinal relationships. In ODM, a time-variant weighting scheme is also introduced to balance the new loss with the standard loss. 
\par 
We evaluate our ODM on retinal images ($C=5$) and endoscopic images ($C=4$) quantitatively and qualitatively. The quantitative evaluations prove that ODM outperforms traditional generative models, such as standard conditional diffusion models and generative adversarial networks~\cite{stylegan2}. As expected, this superiority was prominent in the highest severity classes with fewer training samples.  
\par 


Our main contributions are summarized as follows:
\begin{itemize}
    \item We expand a diffusion model to generate images with ordinal classes.
    \item We show that controlling the ordinal relationship among noise images is useful for controlling the generated images. 
    \item Experimental evaluations with two public medical image datasets confirm our model outperforms conventional generative models in several performance metrics, such as Frechet Inception Distance (FID).
\end{itemize}
\par
\section{Related work} 
\label{sec:rel_work}
\subsection{Diagnostic Models with Ordinal Classes}
Medical image diagnosis models that learn some ordinal relationships have been proposed so far~\cite{med_or,med_rank,csGAN}. Liu et al.~\cite{med_or} propose several models for dealing with the severity classes of Diabetic Retinopathy (DR). They use a simple multi-class classification model, as well as a multi-task classification model.
Kadota et al.~\cite{med_rank} utilize RankNet~\cite{ranknet} for evaluating the severity of UC. In their work, RankNet is trained so that an endoscopic image with a higher severity gives a higher rank score than another image with a lower severity. The ordinal relationship of medical images has been used in classification and ranking tasks but {\em not} image generation tasks.\par

\subsection{Diffusion Models for Medical Images}
Recently, diffusion models have been widely used for medical image generation~\cite{ldm_brain,dalle_derma,miccai_ddpm_v1,miccai_ddpm_v2}. These models are often conditional on age, disease type, etc., to control the quality of generated images. Pinaya et al.~\cite{ldm_brain} propose a diffusion model that generates brain MRIs with conditions on the age and gender of patients. Sagers et al.~\cite{dalle_derma} train a diffusion model to generate photorealistic dermatology images with 16 skin type conditions. Ye et al.~\cite{miccai_ddpm_v1} propose a latent diffusion model~\cite{ldm} that generates histopathology images with class labels. To the authors' best knowledge, this paper is the first attempt to use diffusion models for generating medical images with some ordinal class conditions.
\par

\section{Ordinal diffusion model}
\label{sec:ODM}

\subsection{Standard Diffusion Models~\cite{ddpm}}
\label{ssec:ddpm}
In the standard diffusion models, the forward process is a diffusion process from $t=1$ to $T$, defined as follows:
\begin{equation}
    \bm{x}_t = \sqrt{1-{\beta}_t}\bm{x}_{t-1} + \sqrt{{\beta}_t}\bm{\epsilon}, 
\end{equation}
where $\bm{x}_0$ is a real image data and $\bm{\epsilon}\sim \mathcal{N}(\bm{0},\bm{I})$ is a random noise image. Therefore, $\bm{x}_t$ is a ``diffused'' real image by adding noise $t$ times. $\beta_t$ is the so-called noise scheduler and defined as $\beta_t = (\beta_T-\beta_1)(t-1)/(T-1)+\beta_1$ with hyperparameters $\beta_1$ and $\beta_T$.\par
The backward process is a denoising process from $t=T$ to $1$ and uses a neural network model $\bm{\epsilon}_{\theta}$ with a weight parameter set $\theta$. Given $\bm{x}_t$ and $t$, the trained model is expected to output $\bm{\epsilon}_{\theta}(\bm{x}_t, t)$, which is an estimated noise added to $\bm{x}_{t-1}$. By using the estimated noise and  $\bm{x}_t$, we can infer $\bm{x}_{t-1}$ by the following equation:
\begin{equation}
    \bm{x}_{t-1} = \frac{1}{\sqrt{\alpha_t}}\left(\bm{x}_t - \frac{1-\alpha_t}{\sqrt{1-\bar{\alpha}_t}}\bm{\epsilon}_\theta(\bm{x}_t,t)\right)+\sigma_t{\bm{z}}, 
\end{equation}
where $\alpha_t = 1-\beta_t$, $\bar{\alpha}_t=\alpha_1\alpha_2\cdots\alpha_t$, and $\sigma_t = \sqrt{\beta_t}$. $\bm{z}$ is a random noise from $\mathcal{N}(\bm{0},\bm{I})$.\par
When it is necessary to generate images of a specific class $c\in \mathcal{C}$, a class-conditional model $\bm{\epsilon}_{\theta}(\bm{x}_{t,c}, t, c)$ has been used, where $\bm{x}_{t,c}$ is the diffused image estimated at time $t$ for the class $c$. The training sample $\bm{x}_{0,c}$ from the 
class $c$ should be used in its training phase.\par
As noted above, the model $\bm{\epsilon}_{\theta}$ should estimate the noise $\bm{\epsilon}$ added to $\bm{x}_t$. For this purpose, the model is trained to minimize the following loss function;
\begin{equation}
    \mathcal{L}_{t}^{\mathrm{DM}} = \mathbb{E}_{\bm{x}_{0,c},\bm{\epsilon}}\left[\|\bm{\epsilon}-\bm{\epsilon}_{\theta}(\bm{x}_{t,c},t,c)\|^2\right].
    \label{eq:mse_noise}
\end{equation}

\begin{figure}[t]
    \centering
    \includegraphics[width=\linewidth]{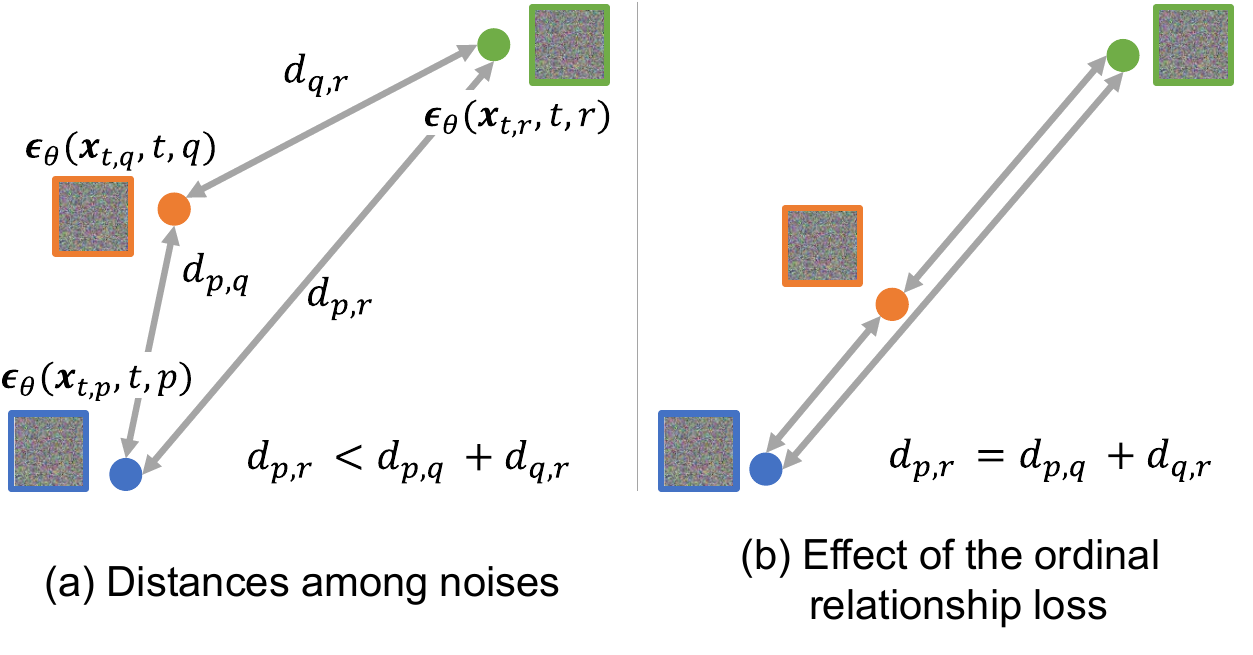}\\[-2mm]
\caption{The effect of the ordinal relationship loss. Class indices $p$, $q$, and $r$ satisfy $1\leq p<q<r\leq C$. Our loss function forces the noise images to lie in a straight line while maintaining their ordinal relationship.}
    \label{fig:ODM}
\end{figure}
\subsection{Training the Model with Ordinal Relationship}
\label{ssec:os_loss}
As noted in Section~\ref{sec:intro}, learning the ordinal relationships between severity classes is important for sharing information from one class with the other. For our task of generating medical images with a diffusion model, we reveal that controlling the estimated {\em noise} images to have some ordinal relationship is an efficient strategy to reflect the ordinal relationships among the classes, as shown in Fig.~\ref{fig:overview}~(b):
\begin{equation}
  \bm{\epsilon}_{\theta}(\bm{x}_{t,p}, t, p)\prec  \bm{\epsilon}_{\theta}(\bm{x}_{t,q},t,q) \prec  \bm{\epsilon}_{\theta}(\bm{x}_{t,r}, t, r), 
\end{equation}
where the class indices $p$, $q$, and $r$ for conditions satisfy $1\leq p<q<r\leq C$.\par
More specifically, we control the estimated noises of individual classes to satisfy the ordinal relationship, as illustrated in Fig.~\ref{fig:ODM}. This control is realized 
by using the following {\em ordinal relationship loss} $\mathcal{L}^{\mathrm{ordinal}}_{t}$ in addition to the standard loss $\mathcal{L}^{\mathrm{DM}}_{t}$ for training the model:
\begin{equation}
    \begin{split} 
        \mathcal{L}^{\mathrm{ordinal}}_{t}
        = \mathbb{E}_{\bm{x}_{0,c},\bm{\epsilon}}[\|&d(\bm{x}_{t,p},\bm{x}_{t,r})-\\&(d(\bm{x}_{t,p},\bm{x}_{t,q})+d(\bm{x}_{t,q},\bm{x}_{t,r})\|^2], 
    \end{split}
    \label{eq:on_loss}
\end{equation}
where the function $d(\cdot,\cdot)$ is the distance between two noises: 
\begin{equation}
    d(\bm{x}_{t,p},\bm{x}_{t,r}) = \|\bm{\epsilon}_{\theta}(\bm{x}_{t,p},t,p)-\bm{\epsilon}_{\theta}(\bm{x}_{t,r},t,r)\|^2.
    \label{eq:d_noise}
\end{equation}
As shown in Fig.~\ref{fig:ODM}~(b), $\mathcal{L}^{\mathrm{ordinal}}_{t}$ achieves its minimum when the three noise images are on a straight line while maintaining their ordinal relationship. In other words, 
the following equation holds with $\exists\alpha\in (0,1)$:
\begin{equation}
  \bm{\epsilon}_{\theta}(\bm{x}_{t,q},t,q) = \alpha\bm{\epsilon}_{\theta}(\bm{x}_{t,p}, t, p)  +(1-\alpha)\bm{\epsilon}_{\theta}(\bm{x}_{t,r}, t, r).
\end{equation}
\par
In ~\cite{ldm}, it is claimed that the denoising process has different characteristics between its beginning steps ($t\sim T$) and ending steps ($t\sim 1$). Specifically, the denoising process focuses on inter-class separability (i.e., semantic variations) when $t\sim T$ and intra-class variations (i.e., perceptual variations) when $t\sim 1$. Since our loss function controls the inter-class ordinal relationship, it should contribute largely at $t\sim T$ and should not at $t\sim 1$. We, therefore, introduce a time-variant weight $\lambda_t = t/T$ and finally the total loss function becomes 
\begin{equation}
    \mathcal{L}^{\mathrm{total}}_t = \mathcal{L}^{\mathrm{DM}}_{t} + \lambda_{t}\mathcal{L}_t^{\mathrm{ordinal}}. 
    \label{eq:loss_ODM}
\end{equation}

\section{Experimental Results}
\label{sec:expr_discuss}

\subsection{Experimental Setup}
\label{ssec:expr_setup}
\subsubsection{Dataset}
\label{ssec:dataset}

We used two medical image datasets with ordinal severity classes for evaluating the proposed model.
\begin{itemize}
    \item EyePACS~\footnote{https://www.kaggle.com/datasets/mariaherrerot/eyepacspreprocess}  contains 35,108 retinal images with five-level ordinal severity classes of Diabetic Retinopathy~(DR). The ordinal classes are No DR (lowest), Mild DR, Moderate DR, Severe DR, and Proliferative DR (highest). The sample sizes for the individual classes are 25,802, 2,438, 5,288, 872, and 708, respectively. 
    \item LIMUC~\cite{limuc} contains 11,276 endoscopic images with the four-level Mayo score, a popular severity class for ulcerative colitis~(UC). According to \cite{uc}, Mayo 0, 1, 2, and 3 correspond to normal or inactive UC, mild UC, moderate UC, and severe UC, respectively. The sample sizes are 6,105, 3,052, 1,254, and 865, respectively. 
\end{itemize}\par
Since the main purpose of the experiment was the observation of the basic performance of ODM, we used the original image space for the denoising process. (We did not employ the lower-dimensional latent space of Stable Diffusion~\cite{ldm}.) Therefore, just like \cite{iddpm}, we used rather small-sized images (64~$\times$~64) in the experiment. For LIMUC, the middle 128~$\times$~128 area was cropped out from the original image ($352\times 288$) with a black border and then resized to 64~$\times$~ 64.
\par 

\begin{figure}[t]
    \centering
    \includegraphics[width=\linewidth]{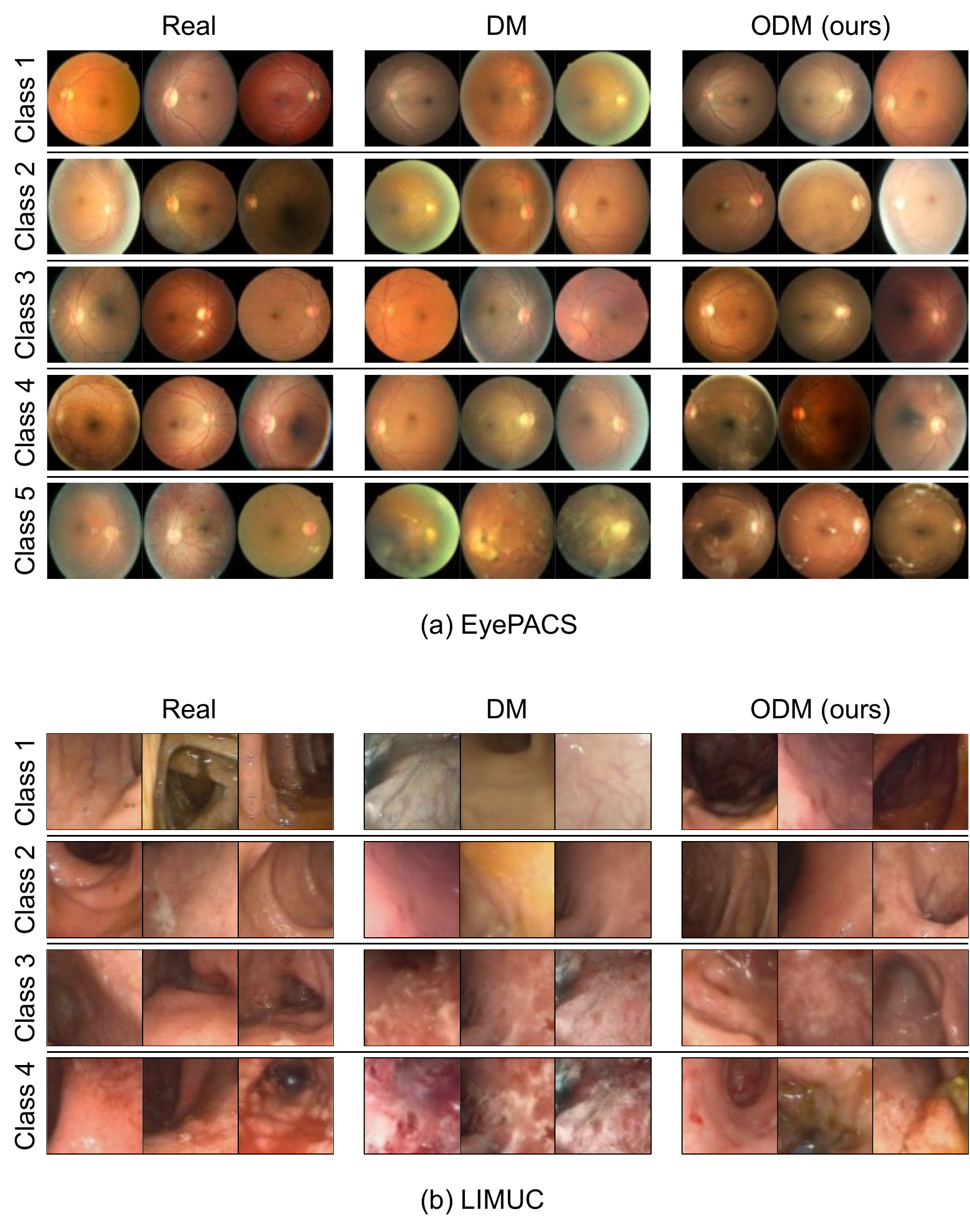}\\[-2mm]
    \caption{Examples of real images and generated images by ODM and the standard diffusion model (DM).}
    \label{fig:gen_imgs}
\end{figure}
\subsubsection{Implimentation Details}
\label{ssec:impliment}
We used the U-Net of Ho et al.~\cite{ddpm} as the denoising model $\bm{\epsilon}_{\theta}$. The U-Net has the residual layer and self-attention to improve the representation performance of the model. We used batch size 32 and 200,000 iterations for the training. We used Adam as the optimizer with the initial learning rate $1.0\times10^{-4}$. We used the same noise scheduler as \cite{ddpm} with ${\beta}_1=10^{-4}$ and ${\beta}_T=0.02$. The total time steps $T$ and DDIM sampling steps~\cite{ddim}  were 1,000 and 100, respectively. 
For the conditional generation, we used classifier-free guidance~\cite{cfg} with the guidance scale of $2$. \par 
\subsubsection{Evaluation Metrics and Comparative Methods}\label{ssec:metric}
We generate 50,000 images from each dataset and then evaluate their quality by FID, a standard metric for evaluating the quality and diversity of AI-generated images. In addition, we use precision and recall defined in \cite{prec_recall}, which evaluates the fidelity of the generated sample distribution to the real sample distribution. More specifically, the precision is defined by the probability that the generated images fall within the real image distribution. The recall is defined by the probability that real images fall within the generated image distribution.
Following tradition, these metrics are not evaluated in the image space but in the feature space with Inception-V3 pre-trained on ImageNet.
\par 
A standard conditional diffusion model (DM) and StyleGAN2~\cite{stylegan2} are used as comparative methods. The former is equivalent to ODM with $\lambda_t \equiv 0$; this means that the result of DM gives an ablation study where the ordinal relationship loss is discarded. For the latter, we used its official implementation. 
\par
\begin{table}[t]
    \centering
    \caption{Quantitative comparison of generated images.  \textcolor{red}{\bf Red}\ indicates the best and \textcolor{blue}{\underline{blue}}\ the second best. In addition to FID, we use precision (``Pre.'') and recall defined in \cite{prec_recall}, to evaluate the overlap between the distributions of real and generated images.}
    \vspace{4mm}
    \begin{tabular}{llccc}
        \hline
         Dataset & Method & FID$\downarrow$ & Pre.$\uparrow$ & Recall $\uparrow$  \\ \hline
         EyePACS & StyleGAN2 & 19.6 & {\bf\color{red} 0.543} & 0.101 \\
                 & DM & {\color{blue}\underline{15.2}} & 0.488 & {\bf\color{red} 0.453} \\
          & ODM (ours) & {\bf\color{red} 11.6} & {\color{blue} \underline{
          
          0.526}} & {\color{blue}\underline{0.443}} \\ \hline
         LIMUC & StyleGAN2 & 60.6 & 0.582 & 0.0811 \\
         & DM & {\color{blue}\underline{37.5}} & {\color{blue}\underline{0.629}} & {\bf\color{red} 0.350}\\
          & ODM (ours) & {\color{red}\bf21.8} & {\color{red}\bf0.702} & {\color{blue}\underline{0.312}} \\ \hline
    \end{tabular}
    \label{tab:performance}
     \color{black}
\end{table}
\subsection{Qualitative Evaluation of Generated Images}
\label{ssec:eval_quality}
Fig.~\ref{fig:gen_imgs} shows generated images by the standard diffusion model (DM), our ODM, and real images. The generated images of DM and ODM do not show significant differences in lower severity classes, which have more training samples. On the other hand, in higher severity classes, where fewer training samples are available, ODM images are more realistic than DM. For example, in the most severe class (Mayo~3, i.e., $c=4$) for LIMUC, ODM could generate endoscopic images with spontaneous bleeding and ulceration, whereas DM generates only bleeding images. 

\subsection{Quantitative Evaluation of Generated Images}
\label{ssec:eval_quantity}
Table~\ref{tab:performance} shows the quantitative comparison result of the 50,000 generated images. The values in the table are measured using the generated images of all classes.  
For FID, the most common metric, ODM largely outperforms the standard diffusion model (DM) and StyleGAN2. This superiority holds in both datasets. For precision and recall (measuring the fidelity and diversity of the distribution), ODM achieved the best or second-best performance. Especially the comparison between ODM and DM (i.e., an ablation where the ordinal relationship loss is discarded)
indicates that the ordinal relationship is useful for generating more realistic images.\par
More importantly, the results prove that controlling the relationship among {\em noises} effectively regulates the ordinal relationship among the final generated images. Although the passive use of the relationship among noises can be found in recent DM research, such as classifier-free guidance~\cite{cfg}, active control of the estimated noises has been an unexplored strategy so far. Noise will be an important medium for precise output control in various DMs. 
\par

\begin{figure}[t]
    \centering
    \includegraphics[width=\linewidth]{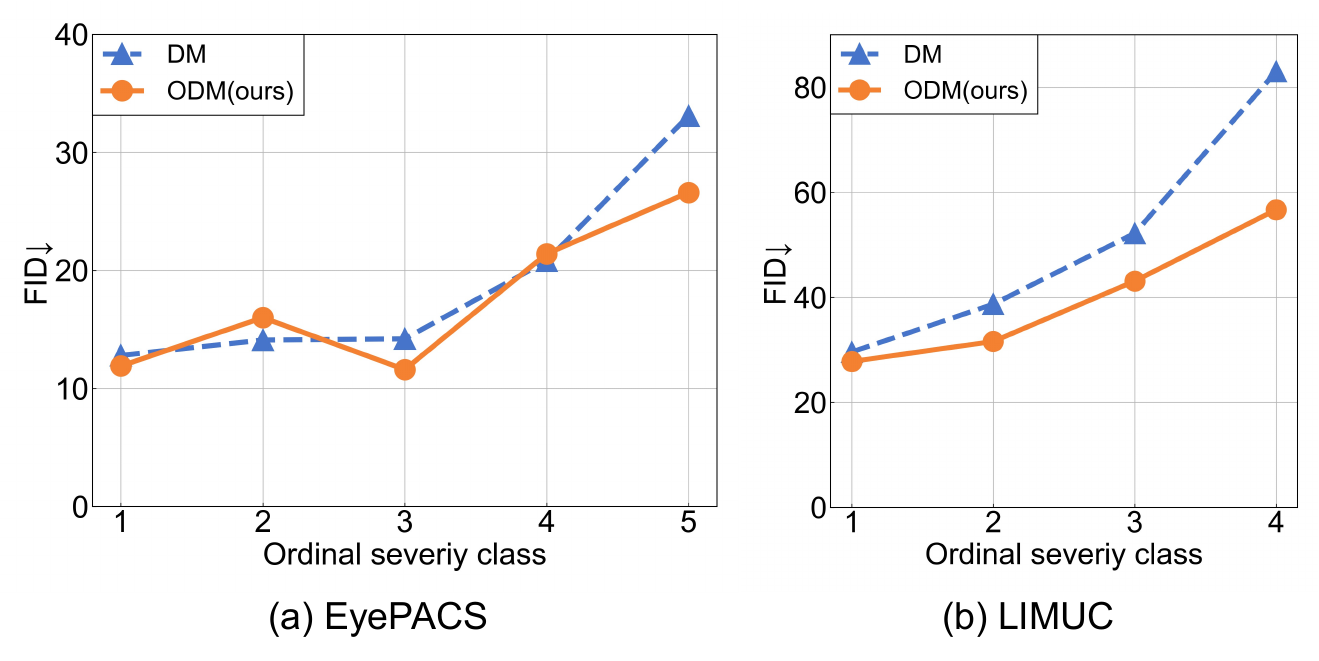}\\
        \caption{Class-wise FID by a standard diffusion model (DM) and ODM.}
    \label{fig:fid}
    \vspace{2.5mm}
\end{figure}

%
Fig.~\ref{fig:fid} plots the class-wise FID of DM and ODM of both datasets. As indicated by the qualitative evaluation of Section~\ref{ssec:eval_quality}, the clear difference between DM and ODM is that ODM achieves better FID than DM, especially at the highest severity classes, where fewer training samples are available. The literature \cite{cbdm} pointed out that the performance of DM degrades by fewer samples; our results indicate that the ordinary relationship relaxes this drawback of DM. This superiority will be beneficial in practice, especially when generating more images from a limited number of training samples for data augmentation.
\par


\section{Conculusion}
\label{sec:conclusion}
We proposed an ordinal diffusion model that learns ordinal relationships among severity classes     
for better medical image generation. The technical highlight of our model is that we control the estimated ``noise'' of different severity classes to satisfy some ordering relationship. Through qualitative and quantitative evaluations of two medical image datasets with severity classes, we confirmed that our model performs better than conventional generative models and can especially generate more realistic images for high-severity classes where sample sizes are limited.\par
Future work is summarized as follows. First, we will modify the ordinal relationship loss to reflect the ordinal class difference ($q-p$ and $r-q$) more strictly to generate more realistic medical images.
Second, we will utilize the generated images in medical image analysis applications. We especially consider, on a classification task, data augmentation for the highest severity classes, where the training samples are often limited. Third, similar to the approach employed in Stable Diffusion, we will perform the denoising process in the latent space, thereby enabling the generation of higher-resolution images.

\newpage
\section{Compliance with Ethical Standards}
This research study was conducted retrospectively using open-access human subject data by \cite{limuc} and Kaggle competition site\footnote{https://www.kaggle.com/datasets/mariaherrerot/eyepacspreprocess}. Ethical approval was not required, as confirmed by the license attached with the open-access data.

\section{Acknowledgments}
This work was supported by JSPS (JP21K18312, JP22H05172, JP22H05173) and SIP (JPJ012425).

\bibliographystyle{IEEEbib}
\bibliography{strings,refs}

\end{document}